\documentclass[a4paper]{article}
\usepackage{sst2016,amssymb,amsmath,epsfig}
\usepackage[hidelinks]{hyperref}
\ninept
\def\reg{{\rm\ooalign{\hfil
     \raise.07ex\hbox{\scriptsize R}\hfil\crcr\mathhexbox20D}}}

\usepackage{csquotes}
\usepackage{tipa}

\newcommand{\SIL}{\textbf{\textipa}} 

\title{Characterisation of speech diversity using self-organising maps}

\makeatletter
\def\name#1{\gdef\@name{#1\\}}
\makeatother
\name{ {\em Tom A. F. Anderson, David M. W. Powers} }
\address{Flinders University\\
{\small \tt tom.anderson@flinders.edu.au, david.powers@flinders.edu.au} }

\begin{document}
\maketitle

\section{Introduction}

We report investigations into speaker classification of larger quantities of unlabelled speech data 
using small sets of manually phonemically annotated speech. The Kohonen speech 
typewriter [1] is a semi-supervised 
method comprised of self-organising maps (SOMs) that achieves low phoneme error rates.
A SOM is a 2D array of cells that learn
vector representations of the data based 
on neighbourhoods.  In this paper, we report a method to 
evaluate pronunciation using multilevel SOMs with
{\SIL/hVd/} single syllable utterances for the study of vowels, 
following [2] (for Australian pronunciation). 

\section{Methodology} 

We used 18 {\SIL/hVd/} words from AusTalk, an Australian speech research corpus collected from
speakers of Australian English [3].
Audio was converted to conventional 39-feature mel-frequency cepstral coefficients 
(MFCCs, log energy, and differential and acceleration 
calculated on 25 ms windows with 10 ms offsets).\footnote{MATLAB scripts are
available from the authors on request. Scripts for MFCCs were modified from
mathworks.com/matlabcentral/fileexchange/32849-htk-mfcc-matlab.}

To investigate the sensitivity of our phonemic model to dialect,
the system was trained using unlabelled speech 
from one of three different groups, General Australian (21 speakers),
educated Melbournian (21 aged 25-34y), 
or speakers of Chinese (14), and labelled using a subset of manually annotated data (phonemes and
start/end points) from three further speakers, as follows: a base 25x25 unit SOM 
was trained using all audio windows from the unlabelled data; three-quarters
of the annotated data was assigned to the units in the base map; the resulting labelled 
map was used for separating the unlabelled input data for submap boosting.
Two approaches to submapping, each with three 20x20 submap SOMs,
were trained on data segmented by the base map by: 
(1)\,three maps for subsets of vowels with
shared confusion in the base map (Kohonen's method); or
(2)\,three maps of either {\SIL/h/}, vowels, or {\SIL/d/} (linguistic).

\section{Results}

Frequently confused phonemes 
(on the first map) were identified using a greedy method 
based on high off-diagonal values in the confusion matrix from the base map:
\begin{itemize}
\item Vowel Group 1: {\SIL/o\textsci/} (hoyd) and {\SIL/o\textlengthmark/} (horde) 
\item Vowel Group 2: {\SIL/e/} (head), {\SIL/e\textlengthmark/} (haired), {\SIL/\textrevepsilon\textlengthmark/} (herd), {\SIL/\textsci/} (hid),
and {\SIL/\textsci\textschwa/} (heared) 
\item Vowel Group 3: {\SIL/\textturna/} (hud), {\SIL/\ae\textopeno/} (howd), {\SIL/\ae\textsci/} (hade), {\SIL/\textturna\textlengthmark/} (hard), {\SIL/\textscripta e/}
(hide), {\SIL/\textopeno/} (hod), and {\SIL/\textschwa\textbaru/} (hode)
\end{itemize}

The vowel error rate, shown in Table 1, was calculated
by comparing the most frequent vowel label outputted for an
audio file with the vowel noted by manual annotators.
Our method of using initial, vowel, or final for submaps was observed to 
be slightly better than Kohonen's method of using submaps to 
disambiguate confused phonemes, and particularly for Chinese, marginally significant.

\begin{table} [thb]
\caption{\label{table1} {\it Vowel Error Rate for single (25x25 classifier),
or multi-SOMs trained
on confused vowels (Vowels)
or initial, vowel, or final (h V d)
(standard error in parentheses) based on 10 CV over word tokens.}}
\vspace{2mm}
\centerline{
\begin{tabular}{|c|ccc|}
\hline
Type & General & Melb. & Chinese \\
\hline  \hline
Single & 0.487(0.062) & 0.600(0.025) & 0.506(0.041)\\
Vowels & 0.059(0.021) & 0.047(0.019) & 0.090(0.016)\\
h V d & 0.057(0.014) & 0.042(0.022) & 0.027(0.015)\\
\hline
\end{tabular}}
\end{table}

\section{Discussion}

The goal of the experiments conducted in this work was to explore
the general capacity of labelled data for revealing characteristics
of dialect in the unlabelled speech of Chinese-background Australians and educated Melbournians 
vs Australians generally. The results are competitive with the original phonetic typewriter results,
demonstrating effectiveness with 39-feature MFCC 
vectors. There is the possibility for improving on these using context, but for the future 
directions of this research, we are more interested in characterising differences between the three groups.

\section{Acknowledgements}
The AusTalk corpus was collected as part of the Big ASC project funded by the Australian Research Council (LE100100211).\footnote{See https://austalk.edu.au/bibliography.html for details.}

\bibliographystyle{IEEEtran}

\end{document}